\documentclass[manuscript,nonacm]{acmart}

\AtBeginDocument{%
  }

\setcopyright{none}
\begin{document}

\title{Towards detecting unanticipated bias in Large Language Models}

\author{Anna Kruspe}
\email{anna.kruspe@hm.edu}
\affiliation{
  \institution{Munich University of Applied Sciences}
  \city{Munich}
  \country{Germany}
}

\renewcommand{\shortauthors}{Kruspe}

\begin{abstract}
Over the last year, Large Language Models (LLMs) like ChatGPT have become widely available and have exhibited fairness issues similar to those in previous machine learning systems. Current research is primarily focused on analyzing and quantifying these biases in training data and their impact on the decisions of these models, alongside developing mitigation strategies. This research largely targets well-known biases related to gender, race, ethnicity, and language. However, it is clear that LLMs are also affected by other, less obvious implicit biases. The complex and often opaque nature of these models makes detecting such biases challenging, yet this is crucial due to their potential negative impact in various applications. In this paper, we explore new avenues for detecting these unanticipated biases in LLMs, focusing specifically on Uncertainty Quantification and Explainable AI methods. These approaches aim to assess the certainty of model decisions and to make the internal decision-making processes of LLMs more transparent, thereby identifying and understanding biases that are not immediately apparent. Through this research, we aim to contribute to the development of fairer and more transparent AI systems.
\end{abstract}



\maketitle

\begin{figure}
\centering
  \includegraphics[width=.9\textwidth]{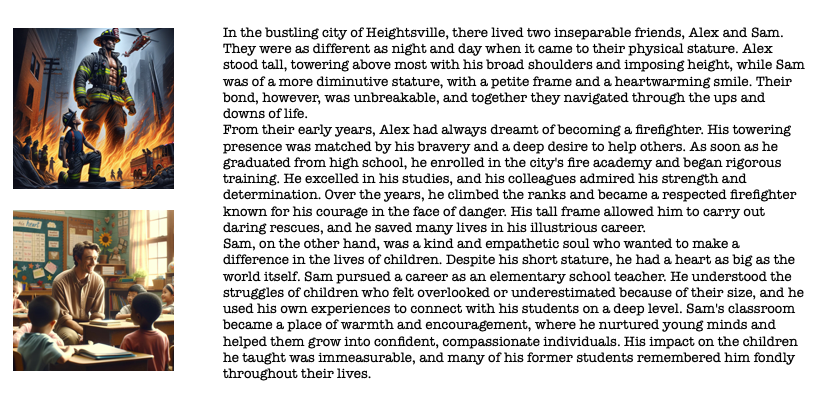}
  \caption{ChatGPT reply for the prompt "Write a story about two friends, one tall and one short, and their careers" and visualizations. The example illustrates a bias that is not often considered, and would not be apparent when prompting directly for stereotypical characteristics of people of different heights. It is easy to see how this implicit bias in GPT-4 could lead to unfair decisions, e.g. in recruiting applications.}
  \label{fig:teaser}
\end{figure}

\section{Introduction}

In recent years, Large Language Models (LLMs) like ChatGPT have gained significant prominence in various sectors, offering innovative solutions to a multitude of Natural Language Processing challenges. Their rise to fame can be attributed to their versatility and ease of use, making them accessible to both experts and laypersons for a wide range of applications, including complex industrial tasks. However, this widespread adoption has overshadowed certain inherent limitations, particularly in real-world applications. One critical issue is the tendency of these models to generate responses that appear naturally intelligent, leading users to overestimate their comprehension abilities. In reality, these models lack genuine understanding, relying instead on patterns learned from extensive training on massive text datasets.

An area of increasing concern is the latent implicit biases within LLMs. Historical language models like word2vec have demonstrated such biases \cite{bolukbasi2016man}, which often go unnoticed until they result in unfair treatment of specific groups. Examples that have come to prominence include gender discrimination in recruitment tools\footnote{\url{https://www.reuters.com/article/us-amazon-com-jobs-automation-insight-idUSKCN1MK08G/}}, advertisement distributions violating the Fair Housing Act\footnote{\url{https://www.cnbc.com/2022/06/21/doj-settles-with-facebook-over-allegedly-discriminatory-housing-ads.html}}, or a chatbot adopting racially biased behavior based on user interactions\footnote{\url{https://arstechnica.com/information-technology/2016/03/tay-the-neo-nazi-millennial-chatbot-gets-autopsied/}}.
While advanced LLMs may mask these biases more subtly, they are not immune to them. Efforts to align LLMs with societal fairness standards, including the development of assistant models and moderation endpoints like those in ChatGPT, have been implemented to mitigate obvious fairness issues \cite{openai2023gpt4}. 

However, these measures do not eliminate biases; they merely provide a post-hoc layer of intervention. This layer might fail, particularly with subtle biases, such as those in unexpected tasks (like programming code or the example in figure \ref{fig:teaser}) or biases not yet thoroughly examined. Intersectionality poses a notable challenge here. Initial studies show that, despite appearances, ChatGPT does display bias \cite{wan2023kelly, singh2023chatgptbiased, urchs2023prevalent}. This issue is more severe in models without these mitigating components, particularly if used for purposes other than dialog systems. The complexity of neural networks, often seen as 'black boxes', exacerbates these problems. Addressing biases solely through training is inadequate, as it restricts the range of training data and does not account for the difficulty of defining 'unbiased' data.

Research in this area is mainly focused on quantifying and mitigating known biases. This paper addresses the challenge of detecting unanticipated biases in LLMs, particularly during the inference stage. We propose employing Uncertainty Quantification (UQ) and Explainable AI (XAI) as tools to unearth biases potentially influencing LLM outputs. These techniques offer insights into whether LLMs are appropriately focusing on a task or if biases are swaying their outputs. Given the extensive computational and data requirements of LLMs, many users depend on pre-trained models with possible minor modifications. Therefore, our emphasis is on post-hoc, model-agnostic methods applicable to any LLM.

The paper is structured to provide an overview of current bias detection methods in LLMs, introduce technical aspects of UQ and XAI, and suggest applications of these techniques in bias detection. We also discuss challenges in evaluation, propose potential bias mitigation strategies, and conclude with the limitations and a summary of our findings. This exploration into Unanticipated Bias Detection (UBD) in LLMs is novel and holds significant promise in addressing a critical aspect of AI fairness and transparency.  With this paper, we aim to bring this topic into greater focus and encourage further research in this area.



\section{Current state of bias and fairness research}\label{sec:sota}

When LLMs started being widely used, it became clear that bias and fairness would be issues to tackle, just like in previous machine learning models trained on human-produced data. As such, there is already a range of research on these topics. We will briefly summarize the main research avenues covered so far. More in-depth overviews are given in \cite{li2023survey, gallegos2023bias}. \cite{mehrabi2022survey} provides a general survey on bias and fairness in Machine Learning, where many of the same statements apply.

\subsection{Bias sources and definitions}
A fundamental question lies in defining what bias and fairness constitute in the context of LLMs. These notions are inherently normative and subjective, varying with the specific task at hand as well as the cultural and contextual milieu. As highlighted by \cite{blodgett2020language}, a common challenge in the technical literature is the lack of clarity and consistency in defining bias. Furthermore, these definitions often overlook the potential harm biased models can inflict, including who is affected and in what ways. Typically, discussions around fairness and bias in LLMs revolve around social groups identified by protected attributes. In this framework, fairness is viewed through two lenses: group fairness, i.e. equitable treatment of groups differing only in their protected attributes; and individual fairness, i.e. treating similar individuals in a comparable manner. Bias, on the other hand, is often understood as deviations from these fairness principles, manifesting in ways that disproportionately and negatively affect certain groups or individuals based on their protected attributes. In NLP, commonly encountered social biases manifest in two primary forms: representational and allocational harms. Representational harms refer to the reinforcement of detrimental stereotypes or negative perceptions about certain social groups. Allocational harms, by contrast, entail the unequal distribution of resources or opportunities among different social groups, leading to disparities in benefits and advantages \cite{gallegos2023bias}.

Related to this is the search for the sources of these biases. The main contributor usually appears to be the training data \cite{mehrabi2022survey}, and indeed most publications focus on detecting biases in data or mitigating it on this basis. However, biases can also stem from the model's design and development, including choices in optimization functions and the handling of training data, which may amplify existing biases. Evaluation practices, such as using non-representative benchmarks and specific performance metrics, contribute to biases by potentially obscuring disparities across different social groups. Additionally, the deployment context and user interface can influence how biases manifest, especially when models are used in settings different from their original intent \cite{suresh2021framework}. 

\subsection{Metrics and evaluation}
Building on the ambiguous definition of bias and fairness, it is not immediately obvious how to detect and evaluate biases numerically.  In the realm of machine learning, traditional statistical measures such as Statistical Parity Difference, Equality of Opportunity, and Disparate Impact have been utilized to assess bias \cite{hort2023bias}. In the specific context of Natural Language Processing (NLP), various tailored metrics have been developed. These metrics include:
\begin{itemize}
\item  Analysis of distances in word or sentence embeddings, e.g. WEAT \cite{caliskan2017semantics} and SEAT \cite{may2019measuring}. CEAT \cite{guo2021detecting} also covers intersectional bias.
\item Evaluation of prediction probabilities, especially when altering protected attributes in tasks like masked token prediction, e.g. DisCo \cite{webster2020measuring}.
\item Examination of text generation system outputs, including LLMs. This involves methods like measuring the distribution of generated co-occurrences, assessing classification predictions made by generative models for different social groups, or comparing outputs against a predefined lexicon of words along with their associated bias scores, e.g. HONEST \cite{nozza2021honest}.
\end{itemize}
These metrics each capture distinct facets of fairness, underscoring the fact that there is no universally applicable solution. Importantly, they all hinge on specific definitions and often require illustrative data examples to effectively quantify the biases they aim to measure.

To facilitate bias evaluation, a number of datasets have been developed. These are usually focused on known biases such as those related to gender, race, ethnic background, etc., and are defined on specific NLP tasks, e.g. Winograd schemas, Coreference Resolution, or Question Answering. Potentially biased samples are hand-labeled and can be used to test systems. A more in-depth analysis is provided in section \ref{sec:evaluation}.

\subsection{Bias mitigation}
A significant portion of current research is dedicated to mitigating biases after their identification in data or models. This mitigation can occur at three key stages in the model's lifecycle: pre-processing, in-processing, and post-processing.
\begin{description}
    \item[Pre-processing] This stage involves addressing bias in the training data before the actual model training begins. The process typically starts with identifying biases within the training dataset. Once these biases are recognized, the data can be modified – for instance, through re-labeling or perturbing – to minimize these biases. Techniques such as re-sampling are also employed to enhance the representation of unbiased data points. Additionally, utilizing data representations that are less prone to bias, or directly de-biasing these representations, are common strategies at this stage.
    \item[In-processing] Most research efforts focus on mitigating bias during the model's training phase. This can be achieved by incorporating regularization methods or constraints within the model to prevent the learning of biased relationships. Adversarial learning is another approach, where an auxiliary model is trained to predict protected attributes from the main model's internal representations, and a min-max strategy is used between the two models. This ideally results in the main model developing representations that do not reflect biases. Other in-processing techniques include training compositional models tailored for each protected group or creating an ensemble of models trained on different data subsets. Adjustments to the learning process itself, such as integrating fairness measures into the hyperparameter optimization process, are also part of this approach.
    \item[Post-processing] In this phase, mitigation efforts focus on models that have already been trained. This involves a detailed analysis of the biases present and their manifestation in the interaction between the model's inputs and outputs. The mitigation might involve adjusting the inputs and outputs to counteract biases, such as replacing certain elements with equivalents from non-protected groups, or modifying the model's weights directly. However, the effectiveness of these techniques diminishes as the complexity of the models and their applications increase.
\end{description}

While these mitigation strategies are promising, it is important to note that they typically guide the models towards reduced bias rather than completely eliminating it. The intricacy of these biases and the models' architecture often means that complete eradication of bias is not yet achievable.

\subsection{Blind spots}
In current research on Large Language Models (LLMs), a significant gap exists: the detection of biases in areas where they are not typically expected. Current studies often focus on well-known biases that have been highlighted due to their clear impact on various tasks. However, this approach overlooks a range of subtle biases that are not immediately obvious, especially in applications that don’t directly involve fairness issues but still process sensitive attributes.

Take the example of using LLMs in medical diagnostics. On the surface, the results might seem accurate and unbiased. But a closer examination might reveal influences from patient attributes that should not affect the diagnosis. These hidden biases, though less apparent, can have serious implications, such as incorrect treatment recommendations \cite{omiye2023large,clusmann2023future}. The situation becomes complex when considering factors like a patient's gender or race. In some diagnostic scenarios, these attributes might be relevant and necessary for accurate diagnosis, while in others, their consideration could introduce harmful biases. Currently, medical practitioners lack the means to discern if and how these factors are factored into the model's decision-making process. This complexity is further magnified when dealing with less obvious or less studied biases, such as those based on physical attributes (like height and weight), demographic factors (such as age or marital status), and their intersectional effects. 

The challenge is compounded by the subjective nature of fairness and the changing definitions of bias over time. What is considered biased in one context may not be seen the same way in another. This evolving understanding of fairness necessitates research not just on known biases, but also on developing systems that can identify and adapt to emerging biases.

This paper aims to address this under-explored area: Unanticipated Bias Detection (UBD) in LLMs. We propose to explore methodologies for identifying these hidden biases. Our goal is to contribute to the development of LLMs that are not only technologically sound but also ethically responsible, capable of adapting to changing societal norms of fairness.

\section{Technical background}\label{sec:technical_background}
We will start by providing technical introductions to two subfields in machine learning that we believe can lead to methods useful for detecting unexpected biases: Uncertainty Quantification (UQ) and Explainable AI (XAI).

\subsection{Uncertainty Quantification (UQ)}
Uncertainty Quantification methods in Machine Learning aim to produce numerical estimates of the degree of certainty a model has about a decision it is making. Such uncertainties can be caused by the task to be solved itself (aleatoric uncertainty) or by deficits in the model's representation of it (epistemic uncertainty). Of particular interest are cases caused by errors in the data foundation, e.g. missing areas of the data space or shifts between the expected data distributions. Therefore, it appears evident that biases would be reflected as uncertainties as well.

There is a variety of approaches for UQ \cite{gawlikowski2023survey}:

\paragraph{Single Deterministic Networks} This approach models both model and data uncertainties within a single deterministic neural network (or a dedicated external model), typically including components that explicitly represent these uncertainties \cite{sensoy2018evidential}. For example, classification networks might have softmax outputs representing data uncertainty, or regression networks might explicitly predict standard deviation. This method integrates uncertainty directly into the model's predictions, offering computational efficiency and less intensity than other methods. However, it may have limitations in capturing the full spectrum of uncertainties and the precision of its uncertainty estimates can be restricted by the network's design. 
    
\paragraph{Bayesian methods}

The core idea of Bayesian Neural Networks (BNNs) is to introduce probabilistic uncertainty into the parameters of neural networks. Unlike traditional neural networks that learn fixed weights, BNNs treat these weights as random variables, learning their probability distributions \cite{gal2016dropout}. This approach provides a principled framework for handling uncertainty, making BNNs particularly valuable in tasks where quantifying uncertainty is as crucial as making predictions.

The implementation of BNNs involves several approaches to approximate the posterior distribution of the weights, given the intractability of direct computation in complex neural network architectures. The three primary methods are Variational Inference, Sampling Approaches, and the Laplace Approximation.

Variational Inference approximates the posterior distribution by optimizing a family of tractable distributions, often Gaussian. The goal is to minimize the difference between the approximated and true posterior, typically measured using Kullback-Leibler divergence. Sampling Approaches, based primarily on Markov Chain Monte Carlo (MCMC) techniques, involve generating samples from the posterior distribution. Laplace Approximation is a simpler method that approximates the posterior distribution as a Gaussian centered around the mode of the posterior. 

Besides their ability to quantify uncertainty, BNNs can also adapt to new data without forgetting previous knowledge, a property known as continual learning, and can inherently avoid overfitting to a certain degree due to their probabilistic nature. The primary challenge is computational complexity, especially in large-scale neural networks, making some BNN approaches computationally expensive and time-consuming. Another issue is the difficulty in specifying meaningful priors, particularly in deep neural networks with a vast number of parameters. This requires careful consideration and domain expertise, as inappropriate priors can significantly skew results.

\paragraph{Ensemble methods} 
The fundamental principle behind ensemble methods is to derive predictions based on the collective output of multiple models, termed ensemble members. This strategy leverages the synergy effects among different models, operating under the premise that a group of decision-makers generally makes better decisions than a single one. Originally conceived to increase robustness, they are now also popular for UQ.

The effectiveness of ensemble methods lies in their ability to perform multi-mode evaluation, as opposed to the single-mode focus of approaches like Bayesian Neural Networks (BNNs). In multi-mode evaluation, ensemble members are encouraged to converge to different local optima, providing a broader exploration of the solution space. This diversity in the ensemble is key to its success and can be achieved through several strategies, including random initialization and data shuffling, bagging and boosting, data augmentation, and diverse network architectures. Uncertainty is then determined by evaluating the variation among members' predictions \cite{lakshminarayanan2017simple}. 

Ensemble methods are appealing due to their straightforward implementation, parallelizable training, and ease of extension. The primary focus is to ensure sufficient diversity among the members, often achieved through simple strategies like random initialization and data augmentation. However, the increasing computational and memory requirements with each added member can be a limiting factor, particularly in large-scale applications or time-critical scenarios. To address these issues, techniques like pruning and distillation have been developed. Pruning reduces the complexity of the ensemble by removing redundant members without significantly impacting performance. Distillation, meanwhile, involves teaching a single network to mimic the behavior of the entire ensemble, thus reducing the overall model count.

\paragraph{Test-time Data Augmentation} 

Test-Time Data Augmentation (TTDA) involves creating multiple variations of each test sample through augmentation techniques, and then using these variations to compute a predictive distribution. This process allows the model to explore different views of the same data, capturing uncertainty more effectively.

The implementation of TTDA is straightforward, requiring no changes to the underlying model and no additional data. However, it is important to use valid augmentations that do not produce data outside the target distribution to avoid altering predictions inaccurately.

Research advancements in TTDA address its limitations. For instance, \cite{shanmugam2020testtime} proposed a learning-based approach that aggregates predictions from each augmented test sample, considering factors like the problem nature, training data size, neural network architecture, and augmentation type. Other researchers, like \cite{molchanov2020greedy, kim2020learning}, have developed methods for selecting the most effective augmentations, either through policy search or loss prediction.

Despite these improvements, determining the impact of different augmentations on uncertainty remains a challenge. It is critical to understand which augmentations capture more uncertainty and how many are necessary for accurately quantifying uncertainties, especially in resource-constrained applications like earth observation. This understanding is key to optimizing TTDA for reliable and effective use in various domains.

\paragraph{Calibration} Another aspect of UQ in Neural Networks lies in calibration, i.e. ensuring that the predicted probability (or uncertainty) produced by the network relates to the actual likelihood of the outcome; e.g. an event predicted by a network with 70\% probability would actually occur in 70\% of real-world cases \cite{kuleshov2018accurate}. This could be directly implemented on model outputs (e.g. softmax values) or applied to the described UQ methods. It is, however, often challenging to transfer such ideas to real-world scenarios due to the variety of uncertainties involved, the mismatch between the training data domain and the unknown actual data domain, and the general inability to model different uncertainties accurately.

\subsection{Explainable AI (XAI)}

Explainability in machine learning is the ability to describe to humans, in understandable terms, how a model makes its decisions or predictions. It is a crucial aspect of AI and machine learning, particularly as these technologies are increasingly used in high-stakes or critical decision-making scenarios. The core idea is to make the behavior of complex models transparent or interpretable. As neural networks are becoming more and more complex, these behaviors become harder and harder to understand; at the same time, highly complex LLMs are increasingly going to be used in real-life scenarios where tracing the decision making process becomes more important to understand. However, the expectation that human users can fully understand such highly complex networks cannot be fulfilled. It is therefore important to define what precisely will help users understand and therefore validate a model's decision.

\cite{zhao2023explainability} states that XAI methods for LLMs can be tackled via two paradigms: Traditional Fine-tuning (as developed on other types of machine learning models), and Prompting. Another distinction lies in whether the methods offer local explanations (i.e. explanations for individual cases/samples/prompts processed by the model) or global explanations (i.e. explaining an aspect of the entire model behavior independent of concrete uses).

\paragraph{Fine-tuning paradigm: Local explanations}
Several classical approaches exist for generating explanations on the local level, each focusing on different aspects of the model.

Feature Attribution-Based Explanation assigns relevance scores to individual input features like words or phrases. It includes Perturbation-based methods that alter input features (e.g., removal, masking) and observe changes in the model's output. The leave-one-out strategy is a classic example, although it may sometimes be misled by feature independence assumptions and overconfident models. As an extension, Example-Based Explanations use specific instances to illustrate model behavior, including adversarial examples that show how small input changes can alter predictions, and counterfactual explanations demonstrating potential outcomes under different input conditions. These examples can be found via perturbation.

Surrogate Models involve replacing complex models with simpler ones (e.g., decision trees, linear models) to approximate the behavior of the original model locally. LIME and SHAP are notable examples \cite{ribeiro2016should, lundberg2017unified}. Decomposition-Based Methods break down the relevance score into linear contributions from each input feature. Techniques like Layer-wise Relevance Propagation (LRP) and Taylor-type Decomposition (DTD) analyze contributions layer by layer or from the final output layer to the input \cite{montavon2015explaining, montavon2019layer}. Gradient-Based Methods analyze how changes in input features affect the model’s output using partial derivatives (gradients). 

Related to this, Attention-Based Explanation Methods focus on the attention mechanism in LLMs, these methods visualize attention patterns and develop function-based techniques to enhance raw attention. The utility of attention in explanations is debated due to limitations in capturing feature importance accurately \cite{jain2019attention}.

\paragraph{Fine-tuning paradigm: Global explanations}

In contrast to local explanations, global explanations for language models attempt to provide comprehensive view of their overall functioning, highlighting how individual components like neurons, hidden layers, and modules contribute to knowledge encoding and linguistic properties.

Probing-Based Explanations involve deciphering the knowledge encapsulated by Large Language Models (LLMs) through self-supervised pre-training. Classifier-based probing, for example, employs a shallow classifier on top of models like BERT to identify specific linguistic properties or reasoning abilities \cite{raffel2020exploring}.
However, the effectiveness of probing classifiers in truly learning syntax, as opposed to just the task, is debated \cite{kunz2020classifier, maudslay2021do}.

Neuron Activation Explanation attempts to identify the importance of individual neurons for performance or linked to specific linguistic properties \cite{bau2018identifying, dalvi2019what}. It examines how shared neurons across different models can imply shared linguistic properties. Methods like greedy Gaussian probing are applied, albeit with challenges in maintaining a balance between accuracy and selectivity \cite{antverg2022pitfalls}. Recent work by OpenAI using GPT-4 has shown promising outcomes in creating natural language explanations for neuron activations \cite{openai2023language}.

Concept-Based Explanation interprets model predictions by mapping inputs to a set of pre-defined concepts, each assigned an importance score. The TCAV (Testing with Concept Activation Vectors) framework exemplifies this, using directional derivatives to quantify the contributions of concepts to model predictions \cite{kim2018interpretability}. Concept-based explanations offer a more human-understandable approach, though they are limited by the need for additional concept-descriptive data and the performance of the concept classifier.

\paragraph{Prompting paradigm}
Rather than generating explanations externally, the Prompting paradigm aims to make models explain their decision process in the same way as they are prompted to perform their tasks. We can differentiate here between approaches for base models (like GPT-4) and for assistant models (those acting on top of the base model to provide dialog capabilities, e.g. ChatGPT). For assistant models, research is mostly focused on alignment \cite{zhou2023lima} and Uncertainty Quantification \cite{xiong2023can} (see section \ref{sec:uncertainty}).

For base models, novel modes of prompting are being developed which elicit emerging properties of the models themselves without requiring explicit retraining or fine-tuning. In-Context Learning (ICL) and Chain-of-Thought prompting are two such strategies which aim to both direct LLMs more explicitly towards users' goals and make their reasoning more transparent. This means that such approaches can both serve to make model outputs better explainable, but there is also work on explaining the effects of the strategies themselves.

In-Context Learning describes the ability of a model to learn and adapt to new tasks or understand new information based on the context provided within the input data, without requiring explicit retraining or fine-tuning. One study uses the SST-2 sentiment analysis benchmark to explain how ICL works in LLMs, finding that the impact of different demonstration types varies with model scale and task type \cite{li2023towards}. Another study investigates whether ICL in large models is enabled by semantic priors or if they learn input-label mappings from provided examples (few-shot learning) \cite{wei2023larger}. Experiments indicate that larger models can learn arbitrary input-label mappings, challenging the view that ICL is solely driven by leveraging priors.

CoT prompting functions by providing step-by-step reasoning examples for certain tasks to a model as few-shot examples to cause it to explain its own reasoning, a method that has significantly improved performance in arithmetic, symbolic, and common-sense reasoning tasks \cite{wei2022chain}. To understand this effect better, \cite{wu2023analyzing} proposes calculating token-wise saliency scores on the input. Other research focuses on perturbing CoT demonstrations in few-shot prompts to determine critical aspects for generating high-performing explanations \cite{madaan2022text, wang2022towards}.



\section{Detecting bias via uncertainty}\label{sec:uncertainty}

As stated above, it appears evident that biases in LLMs would lead to higher uncertainty on their part. The assumption here is that biased data would only make up part of the training for certain concepts, and even partially be counteracted by text data that makes these biases explicit and argues against them. To our knowledge, UQ has not so far been employed for UBD. However, there is existing research confirming the link between biases and uncertainty in other types of Neural Networks: \cite{heuss2023predictive} proposes bias mitigation via ensemble uncertainty in web-search ranking tasks, \cite{stone2023implicit} demonstrates the relationship between BNN bias and uncertainties and its use for mitigation in image sharpening, and \cite{napoles2022fuzzyrough} argues that bias can be expressed as fuzzy-rough uncertainty in pattern classification.

A major roadblock in performing research on LLMs lies in their computational complexity. Training them from scratch or even finetuning them requires vast amounts of data and computational power that are often not available to researchers. Moreover, the ability to detect biases on models that are already in use, and even proprietary ones, would have a much wider range of applications than limiting detection approaches to dedicated LLMs. We will therefore focus on approaches that can be applied post-hoc, particularly model-agnostic methods.

Several classical approaches, such as Bayesian Neural Networks \cite{gal2016dropout}, involve alterations to the fundamental models, which is not viable for these reasons. The Single Deterministic Model approach would also require the model to be trained to predict its uncertainty from the start. Alternatively, a secondary model could be trained to predict a LLM's uncertainties, but it is not straightforward how to obtain ground-truth uncertainties for this training. We therefore propose using the remaining UQ approaches: Test Time Augmentation, Ensemble Methods, and Verbal Uncertainty.

\paragraph{Test-Time Data Augmentation} As described above, TTDA involves intelligently altering the model inputs in a way that should still lead to the same outcome, allowing for the testing of multiple variations of a query. Uncertainty is then determined via analyzing the variation of results. Data augmentation in the text domain has become a more prominent research topic in recent years \cite{Shorten2021TextDataAugmentation}. For LLMs, TTDA may involve manipulating the prompt or context in semantically useful ways, e.g. replacing words with synonyms or automatically paraphrasing sentences. This also poses interesting questions with regards to protected attributes or populations within the data; e.g. should words be replaced not just with synonyms, but also with antonyms in some cases (i.e. counterfactuals like ``man'' vs. ``woman'') \cite{Zmigrod2019CounterfactualDA}, and how could such ``bias-blind'' augmentations be implemented \cite{mouli2022bias}. 

\paragraph{Ensemble Methods} Here, the same queries are tested using different models, and their results are compared \cite{lakshminarayanan2017simple}. When working with LLMs, it would be possible to utilize a set of pre-trained models instead of a singular one. For UBD, it is fairly unlikely that models trained on different data and with different architectures would all possess the same bias, leading to higher uncertainty. Frameworks such as \textit{GPT4All} \cite{anand2023gpt4all} unify model interfaces, and would therefore simplify accessing many responses at once. However, this of course necessitates a multiple of the original computational power, and takes model selection out of the users' hands, or requires them to set up several models. A more straightforward case lies in performing multiple iterations with the same model (Self-Consistency) by exploiting random factors within the model itself (e.g. Temperature Sampling) \cite{wang2022selfconsistency}. This still increases computational complexity but removes the need for multiple models. Strategies include prompting the model in the same way several times, or employing different prompting strategies for the same task \cite{xiong2023can}. The assumption for our task here is that a bias would not surface for every iteration.

\paragraph{Verbal Uncertainty} Finally, models could be directly prompted to provide an estimate of their uncertainty. Unfortunately, experiments show that models are often overconfident, and special methods are needed to balance this effect \cite{xiong2023can}. Such strategies include multi-step Verbalized confidence, i.e. splitting the task into multiple steps and prompting for confidence in each one, and Top-K verbalized confidence, i.e. prompting for the Top-K guesses and their respective likelihoods. With regards to UBD, the first approach may cause the model to recognize biased influences more easily (and for users to pinpoint them better as well), while the second approach functions similarly to the ensemble methods described above. Chain-of-Thought-based verbalized confidence, i.e. prompting the model to think step by step, is another possibility, but may decrease the quality of the response. Combining verbal uncertainty elicitation with the ensemble methods from above is also possible.

\section{Detecting bias via XAI}\label{sec:XAI}

XAI is another promising research direction for detecting unanticipated biases in LLMs. While not providing a fully automated detection out of the box, XAI methods uncover the major factors contributing to a model's output, and therefore enable users to spot influences that should not lead to certain decisions (e.g. protected attributes). Downstream, users can then adapt their prompting strategies or data provided to the model accordingly. The method can then also be utilized for individual use cases to automatically detect biases previously found via exploration with it. As in Uncertainty Quantification, there is no research on using XAI for UBD yet to our knowledge, but literature confirms its feasibility for bias detection in other Neural Networks. This includes \cite{sousa2021explainable}, where surrogate model explainability approaches are used to detect biases in image classification for COVID CT scans, and \cite{wachter2017counterfactual}, which proposes methods for ensuring GDPR-required decision transparency via Counterfactual explanations.

The same qualifications stated above about preferring post-hoc approaches also apply to XAI. We will therefore not look into methods involving an analysis of activations, gradients, attention, etc., or those that require training new models, e.g. in order to capture explanatory concepts. This means that most of the approaches for global explanations from section \ref{sec:XAI} are not feasible, as well as some approaches for local explanations.
Instead, our focus will be on Perturbation-based approaches, Surrogate models, and Prompting.

\paragraph{Perturbation-based Approaches} As described above, these approaches involve the deliberate manipulation of model inputs to analyze their impact on the output \cite{wu2020perturbed, li2016visualizing}. These manipulations can include masking, deleting, swapping, inserting, or replacing tokens. This is related to Test-Time Data Augmentation 
and can be performed with two goals: Adversarial and Counterfactual perturbations.
Counterfactuals describe semantically meaningful changes in the input data  that are expected to alter the output \cite{wu2021polyjuice}. This is particularly interesting for UBD, where we could use large sets of manipulations targeted towards protected attributes. In contrast, adversarial changes are the most minor ones detected that lead to a significant impact on the outcome \cite{jin2020bert}. This too could help uncover unanticipated bias by elucidating input factors that were not expected to contribute.

\paragraph{Surrogate Models} Building on those simpler methods, it is also possible to train simpler surrogate models to understand more complex relationships between inputs and outputs. The advantage here is that various features in the input (prompts, context) may not individually change the output significantly, but would in combination lead to unexpected results. Classical methods include LIME \cite{ribeiro2016why} and SHAP \cite{lundberg2017unified}. For LLMs, this could be implemented by performing a range of perturbations on an input or a whole class of tasks, and then e.g. fine-tuning a smaller LM with those. An obvious application for UBD are intersectional problems. As we suspect that LLMs contain subtle biases that go beyond simple changes when protected attributes are changed, such approaches could be extremely helpful.

\paragraph{Prompting} Directly prompting models to explain their decisions can yield unpredictable results, varying in accuracy and susceptibility to hallucinations (especially when considering desirability). As described above, In-Context Learning and Chain-of-Thought prompting are promising directions here. ICL inputs can be perturbed as described above to determine their saliency, revealing what they contribute to the model. For UBD, this means that we can see whether the model focuses on any bias issues in the context, or more explicitly, whether fairness-related instructions in the context are considered. CoT prompting evokes a step-by-step explanation from the model, which can reveal biased reasoning more explicitly than only considering individual feature impact.

\section{Evaluation}\label{sec:evaluation}


Evaluation of bias detection in (L)LMs is still a relatively new research direction. Fortunately, there is already a range of data sets in existence that provide examples of potentially bias-prone tasks and facilitate numeric evaluations. Such tasks include Counterfactual Inputs on the token or sentence level. In the first case, the goal is usually Masked Token prediction (i.e. Winograd schemas - which token fits best into a gap). On the sentence level, tasks include Coreference Resolution, Entailment, and Intra- and Intersentence likelihood predictions. For evaluation, samples with biased and unbiased solutions are provided, and a model's bias and fairness are evaluated with regards to how well its responses align in each case. For LLMs specifically, there are data sets which mainly focus on prompts for sentence completion, and Question Answering \cite{gallegos2023bias}.

There is some criticism of the construction of data sets in this way, as the samples can only capture a narrow definition of bias, and may reflect the creators' own biases and blind spots. Moreover, it is not always obvious what types of bias the samples cover, and whether results would generalize to other populations \cite{blodgett2021stereotyping}.

When transferring these ideas to UBD, more issues arise. First of all, data sets so far strongly focus on a small set of frequently observed biases such as gender, race, and religion. Second, the contained samples make the bias issue fairly obvious (e.g. when feeding tasks from \textit{StereoSet} \cite{nadeem2021stereoset} to ChatGPT, ChatGPT immediately recognizes that the prompt is targeting a bias problem). We expect LLMs to have biases on a much more subtle level.

Unanticipated biases are, by definition, hard to evaluate. The only currently available option lies in utilizing a very wide range of the described data sets to uncover as much as possible about the model. For future research, a data set of prompts that could be influenced by more subtle biases in the model would be immensely useful; however, this is once again fairly subjective and model-dependent. As described in the next section, results of the approaches proposed by us can serve to make unanticipated biases more transparent to users while still requiring their own feedback and decisions. In the long run, prompts, undesirable responses and their desired alternatives found by users in this way could be collected to construct a large data set for deeper analysis of the issue and targeted improvements. 

A different evaluation question lies in how to evaluate the proposed methods with regards to UBD themselves. Here, using the existing data sets would once again be a good starting point. Experiments could determine whether UQ and XAI deliver useful results on these materials, and where their strengths and weaknesses lie. A parallel strategy could consist of creating main prompts with potential bias, and attaching auxiliary ones to cause the model to focus on these biases (e.g. act as a specific persona) in its response. For evaluation, the auxiliary prompt would be hidden. Finally, as explained above, models themselves can be prompted to provide explanations or statements about their uncertainty, but these are not reliable due to overconfidence or hallucinations. Future research could compare those with the results determined by other, external UQ and XAI approaches to see how they align, and potentially improve the model's own statements.

\section{Proposed mitigation methods}\label{sec:mitigation}

\begin{figure}
    \centering
    \includegraphics[width=0.7\linewidth]{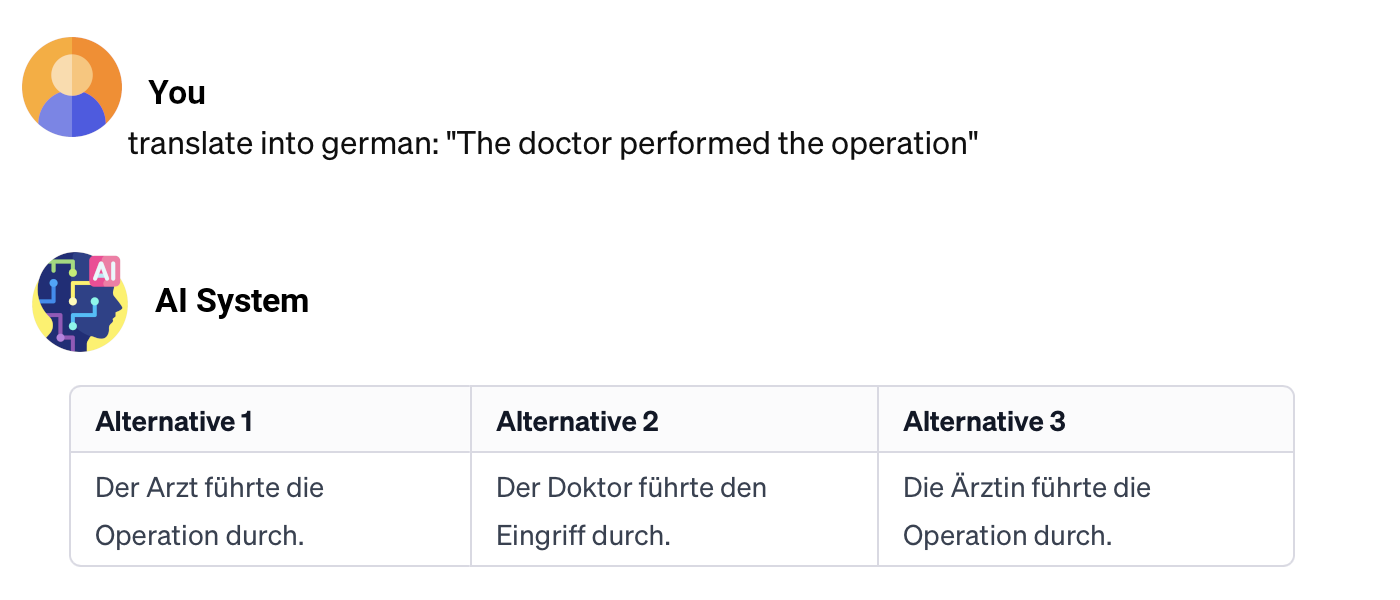}
    \caption{A mockup of a potential uncertainty result: Offering multiple response alternatives, in this case for the task of translating into a language with gendered inflections. [Icons: flaticon.com]}
    \label{fig:mockup_alternatives}
\end{figure}

\begin{figure}
    \centering
    \includegraphics[width=1\linewidth]{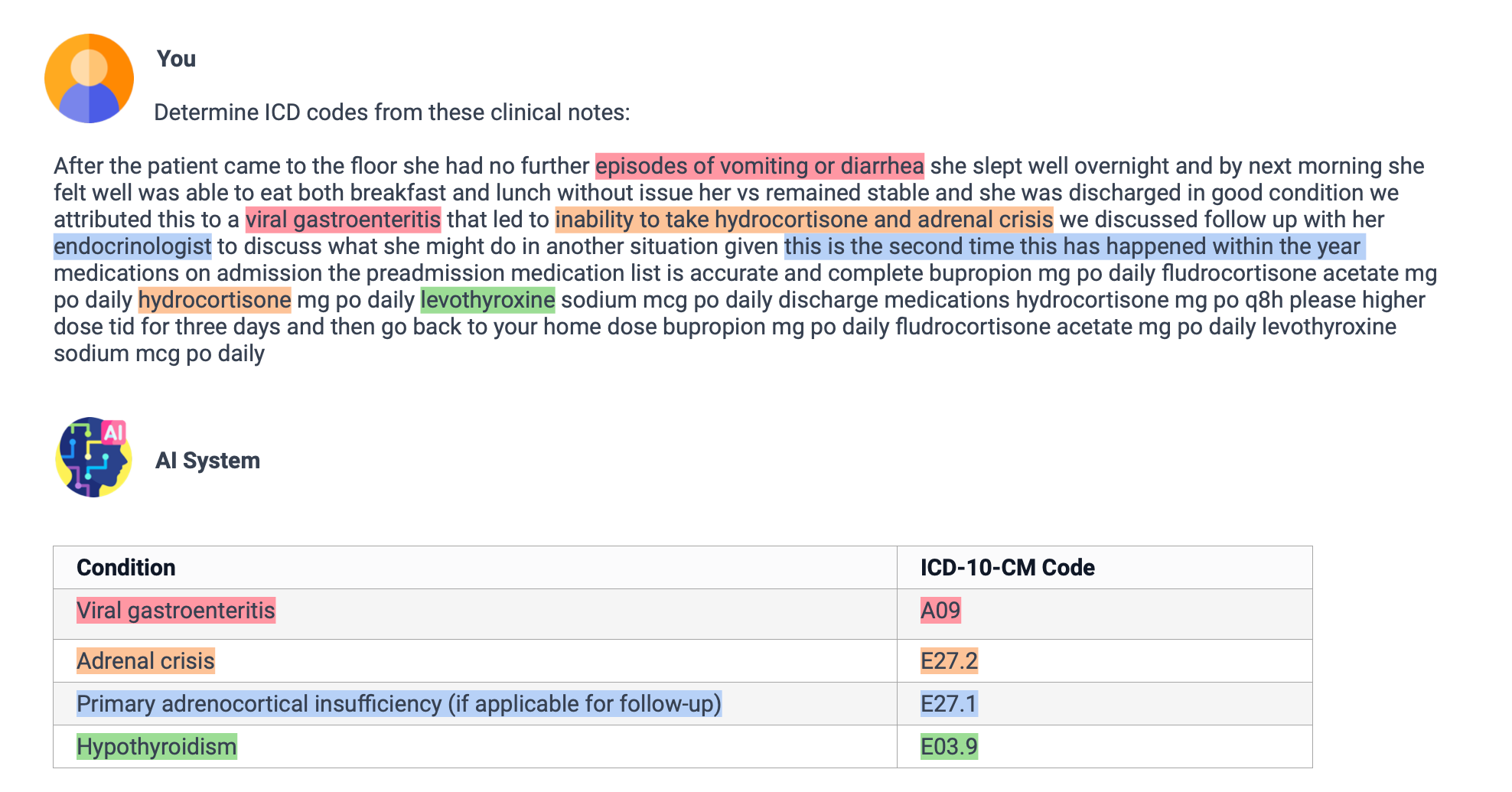}
    \caption{A mockup of a potential explainability result: Demonstrating influence factors in the input. Medical example from \cite{nguyen}. [Icons: flaticon.com]}
    \label{fig:mockup_XAI}
\end{figure}

In both suggested paths, the focus lies on local approaches that lead to statements about the model's behavior in response to a specific query, as opposed to those aiming for a general statement about the model. The search for unexpected biases would not allow for such general statements. Moreover, concentrating on local approaches allows for a more detailed and tailored analysis of how models behave in response to specific queries. This is crucial for understanding nuanced interactions and behaviors that might not be apparent in broader, general analyses. By focusing on specific use cases or queries, the research can develop more customized solutions and strategies that are directly relevant to the user's needs. However, it would be conceivable to systematically apply the developed methods to larger datasets to make assessments about the model for certain biases or for specific use cases. The potential to scale the developed methods to larger datasets offers the opportunity to validate the findings across a broader spectrum of data and use cases.

The developed approaches do not allow for fully automated detection of biases. As stated above, biases in general are too subjective for this purpose, and it would be necessary to define in advance which factors should be considered biases and which should not for each use case. Instead, these approaches offer users a valuable foundation for recognizing unanticipated biases themselves. By providing users with the tools and knowledge to recognize biases, the research empowers them to make informed decisions and adjustments based on their understanding of the model's behavior. 

There are several practical ways to implement tools to help users better understand complex model behaviors and biases. One suggestion lies in adding visual wrappers to existing systems that make the results obtained via UQ and XAI transparent. For example, such wrappers could display response alternatives obtained via UQ, or highlight which parts of the input were detected to be relevant for the response via XAI. Mockups for these ideas are shown in figures \ref{fig:mockup_alternatives} and \ref{fig:mockup_XAI}.

In the long term, users can also learn where the weaknesses of the models lie in relation to their specific use cases and how these might already be compensated for on their side. Besides teaching users to be aware of a model's deficiencies and possibly adapting their prompts and workflow accordingly, this process also enables users to contribute to model improvement. User feedback, i.e., which result was ultimately desired and which was not, can be collected and prospectively used to fine-tune models for increased fairness. This feedback loop creates an iterative process where models are continually refined based on user interactions and needs, leading to more equitable and effective AI systems.

The strategic focus on local approaches, systematic application to larger datasets, recognition of biases through non-automated means, development of visualization tools, long-term learning and adaptation, and collecting user feedback for model fine-tuning represents a comprehensive and user-centered methodology. This approach not only enhances the understanding of AI models but also empowers users to actively participate in the detection and mitigation of biases, leading to more fair and transparent AI applications.

\section{Limitations}\label{sec:limitations}

While we believe that research into UBD in LLMs is highly necessary and timely, we recognize that there are limitations to the proposed method, and potentially all research in this field. First of all, any data generated by humans holds the potential of carrying bias, even if it may not be immediately obvious. As such, it appears impossible to obtain completely bias-free data and models. This is a particular issue in LLMs due to their requirements for huge amounts of training data, which makes curating it infeasible, and does not allow discarding large swathes of the data. The subjectiveness of bias and fairness aggravates this issue. As a special case, there are use cases where bias is desired (e.g. selecting the most intelligent candidate for a job).

Another problem lies in the closed sources of many popular LLMs. While we focused here on post-hoc methods, future developments may show that tackling the problem at a level closer to the model itself is more salient, whether it be analyzing these models in more depth at the computational level or finetuning them for higher bias sensitivity. Even with the methods suggested by us, access to the inner workings could contribute to evaluation.

While UQ and XAI are promising techniques for unanticipated bias detection, there is currently no way to disentangle their results with regards to other influences. In the case of UQ, other factors than biases may contribute to a model's uncertainty, e.g. requests operating beyond the model's capabilities such as modalities not covered by the model. For XAI, we may find unexpected explanation factors that are not caused by bias. It could, however, be argued that any conflict within the model's world representation leading to confusion reflects a sort of bias. On the flip side, it is entirely plausible that the suggested approaches will not uncover any bias deemed socially unacceptable. A possible reason for this could be a strong overrepresentation of the bias in the training data and model.

As described above, UQ and XAI by themselves cannot fully automatically detect biases, but can only offer researchers and users clues where those may occur. This leaves the responsibility for the actual detection in users' hands. This is not necessarily a bad aspect as users will most likely be the experts on what would constitute a bias problem for their task, but it requires high attentiveness on their part. Moreover, humans of course also carry their own prejudices, and may therefore not be able to spot such issues even when additional information is provided.

\section{Summary}\label{sec:summary}
This paper addresses the critical issue of unanticipated biases in Large Language Models (LLMs). As the use of LLMs in various domains increases, it becomes imperative to understand and mitigate the potential biases these models may inadvertently perpetuate. This research explores innovative approaches to identify biases that are not immediately obvious, contributing significantly to the field of Fairness, Accountability, and Transparency in AI.

We presented a comprehensive analysis of potential methods for detecting unanticipated biases in LLMs. Our focus on Uncertainty Quantification (UQ) and Explainable AI (XAI) methods provides a novel perspective on understanding and interpreting the complex behaviors of LLMs. The proposed strategies aim to enable users to recognize biases in specific instances, thereby promoting a more informed and responsible use of these powerful models.

Our approach emphasizes the importance of local explanations, offering users practical tools to understand how LLMs arrive at particular decisions. This user-centric methodology not only enhances the transparency of AI models but also empowers users to actively participate in the detection and mitigation of biases. The proposed visualization tools and user feedback mechanisms are designed to facilitate this process, making it more accessible and effective.

However, we acknowledge the limitations of our approach. The inherent subjectivity of bias, the closed-source nature of many popular LLMs, and the challenge of disentangling other influences in UQ and XAI results pose significant challenges. Moreover, the responsibility for bias detection largely remains with the users, demanding their vigilance and critical evaluation.

In summary, this paper contributes to the ongoing discourse on bias in AI by providing new insights and methods for unanticipated bias detection in LLMs. Our research highlights the need for continuous evaluation and adaptation of AI models, emphasizing the role of users in shaping fair and equitable AI systems. As the field evolves, we hope our work will inspire further research and development in creating more transparent, accountable, and unbiased AI technologies.


\newpage
\bibliographystyle{ACM-Reference-Format}
\bibliography{sources}


\begin{thebibliography}{66}


\ifx \showCODEN    \undefined \def \showCODEN     #1{\unskip}     \fi
\ifx \showDOI      \undefined \def \showDOI       #1{#1}\fi
\ifx \showISBNx    \undefined \def \showISBNx     #1{\unskip}     \fi
\ifx \showISBNxiii \undefined \def \showISBNxiii  #1{\unskip}     \fi
\ifx \showISSN     \undefined \def \showISSN      #1{\unskip}     \fi
\ifx \showLCCN     \undefined \def \showLCCN      #1{\unskip}     \fi
\ifx \shownote     \undefined \def \shownote      #1{#1}          \fi
\ifx \showarticletitle \undefined \def \showarticletitle #1{#1}   \fi
\ifx \showURL      \undefined \def \showURL       {\relax}        \fi
\providecommand\bibfield[2]{#2}
\providecommand\bibinfo[2]{#2}
\providecommand\natexlab[1]{#1}
\providecommand\showeprint[2][]{arXiv:#2}

\bibitem[Anand et~al\mbox{.}(2023)]%
        {anand2023gpt4all}
\bibfield{author}{\bibinfo{person}{Yuvanesh Anand}, \bibinfo{person}{Zach Nussbaum}, \bibinfo{person}{Brandon Duderstadt}, \bibinfo{person}{Benjamin Schmidt}, {and} \bibinfo{person}{Andriy Mulyar}.} \bibinfo{year}{2023}\natexlab{}.
\newblock \bibinfo{title}{GPT4All: Training an Assistant-style Chatbot with Large Scale Data Distillation from GPT-3.5-Turbo}.
\newblock
\newblock
\urldef\tempurl%
\url{https://github.com/nomic-ai/gpt4all}
\showURL{%
\tempurl}


\bibitem[Antverg and Belinkov(2022)]%
        {antverg2022pitfalls}
\bibfield{author}{\bibinfo{person}{Omer Antverg} {and} \bibinfo{person}{Yonatan Belinkov}.} \bibinfo{year}{2022}\natexlab{}.
\newblock \showarticletitle{On the Pitfalls of Analyzing Individual Neurons in Language Models}.
\newblock \bibinfo{journal}{\emph{arXiv preprint arXiv:2110.07483}} (\bibinfo{date}{August} \bibinfo{year}{2022}).
\newblock
\urldef\tempurl%
\url{http://arxiv.org/abs/2110.07483}
\showURL{%
\tempurl}


\bibitem[Bau et~al\mbox{.}(2018)]%
        {bau2018identifying}
\bibfield{author}{\bibinfo{person}{Anthony Bau}, \bibinfo{person}{Yonatan Belinkov}, \bibinfo{person}{Hassan Sajjad}, \bibinfo{person}{Nadir Durrani}, \bibinfo{person}{Fahim Dalvi}, {and} \bibinfo{person}{James Glass}.} \bibinfo{year}{2018}\natexlab{}.
\newblock \showarticletitle{Identifying and controlling important neurons in neural machine translation}.
\newblock \bibinfo{journal}{\emph{arXiv preprint arXiv:1811.01157}} (\bibinfo{year}{2018}).
\newblock


\bibitem[Blodgett et~al\mbox{.}(2020)]%
        {blodgett2020language}
\bibfield{author}{\bibinfo{person}{Su~Lin Blodgett}, \bibinfo{person}{Solon Barocas}, \bibinfo{person}{Hal Daum{\'e}~III}, {and} \bibinfo{person}{Hanna Wallach}.} \bibinfo{year}{2020}\natexlab{}.
\newblock \showarticletitle{Language (technology) is power: A critical survey of “bias” in NLP}. In \bibinfo{booktitle}{\emph{Proceedings of the 58th Annual Meeting of the Association for Computational Linguistics}}. \bibinfo{publisher}{Association for Computational Linguistics}, \bibinfo{address}{Online}, \bibinfo{pages}{5454--5476}.
\newblock
\urldef\tempurl%
\url{https://doi.org/10.18653/v1/2020.acl-main.485}
\showDOI{\tempurl}


\bibitem[Blodgett et~al\mbox{.}(2021)]%
        {blodgett2021stereotyping}
\bibfield{author}{\bibinfo{person}{Su~Lin Blodgett}, \bibinfo{person}{Gilsinia Lopez}, \bibinfo{person}{Alexandra Olteanu}, \bibinfo{person}{Robert Sim}, {and} \bibinfo{person}{Hanna Wallach}.} \bibinfo{year}{2021}\natexlab{}.
\newblock \showarticletitle{Stereotyping {N}orwegian Salmon: An Inventory of Pitfalls in Fairness Benchmark Datasets}. In \bibinfo{booktitle}{\emph{Proceedings of the 59th Annual Meeting of the Association for Computational Linguistics and the 11th International Joint Conference on Natural Language Processing (Volume 1: Long Papers)}}, \bibfield{editor}{\bibinfo{person}{Chengqing Zong}, \bibinfo{person}{Fei Xia}, \bibinfo{person}{Wenjie Li}, {and} \bibinfo{person}{Roberto Navigli}} (Eds.). \bibinfo{publisher}{Association for Computational Linguistics}, \bibinfo{address}{Online}, \bibinfo{pages}{1004--1015}.
\newblock
\urldef\tempurl%
\url{https://doi.org/10.18653/v1/2021.acl-long.81}
\showDOI{\tempurl}


\bibitem[Bolukbasi et~al\mbox{.}(2016)]%
        {bolukbasi2016man}
\bibfield{author}{\bibinfo{person}{Tolga Bolukbasi}, \bibinfo{person}{Kai-Wei Chang}, \bibinfo{person}{James Zou}, \bibinfo{person}{Venkatesh Saligrama}, {and} \bibinfo{person}{Adam Kalai}.} \bibinfo{year}{2016}\natexlab{}.
\newblock \bibinfo{title}{Man is to Computer Programmer as Woman is to Homemaker? Debiasing Word Embeddings}.
\newblock
\newblock
\showeprint[arxiv]{1607.06520}~[cs.CL]


\bibitem[Caliskan et~al\mbox{.}(2017)]%
        {caliskan2017semantics}
\bibfield{author}{\bibinfo{person}{Aylin Caliskan}, \bibinfo{person}{Joanna~J Bryson}, {and} \bibinfo{person}{Arvind Narayanan}.} \bibinfo{year}{2017}\natexlab{}.
\newblock \showarticletitle{Semantics derived automatically from language corpora contain human-like biases}.
\newblock \bibinfo{journal}{\emph{Science}} \bibinfo{volume}{356}, \bibinfo{number}{6334} (\bibinfo{year}{2017}), \bibinfo{pages}{183--186}.
\newblock
\urldef\tempurl%
\url{https://doi.org/10.1126/science.aal4230}
\showDOI{\tempurl}


\bibitem[Clusmann et~al\mbox{.}(2023)]%
        {clusmann2023future}
\bibfield{author}{\bibinfo{person}{Jan Clusmann}, \bibinfo{person}{Fiona~R. Kolbinger}, \bibinfo{person}{Hannah~Sophie Muti}, \bibinfo{person}{Zunamys~I. Carrero}, \bibinfo{person}{Jan-Niklas Eckardt}, \bibinfo{person}{Narmin~Ghaffari Laleh}, \bibinfo{person}{Chiara Maria~Lavinia L{\"o}ffler}, \bibinfo{person}{Sophie-Caroline Schwarzkopf}, \bibinfo{person}{Michaela Unger}, \bibinfo{person}{Gregory~P. Veldhuizen}, \bibinfo{person}{Sophia~J. Wagner}, {and} \bibinfo{person}{Jakob~Nikolas Kather}.} \bibinfo{year}{2023}\natexlab{}.
\newblock \showarticletitle{The future landscape of large language models in medicine}.
\newblock \bibinfo{journal}{\emph{Communications Medicine (London)}} \bibinfo{volume}{3}, \bibinfo{number}{1} (\bibinfo{year}{2023}), \bibinfo{pages}{141}.
\newblock
\urldef\tempurl%
\url{https://doi.org/10.1038/s43856-023-00370-1}
\showDOI{\tempurl}


\bibitem[Dalvi et~al\mbox{.}(2019)]%
        {dalvi2019what}
\bibfield{author}{\bibinfo{person}{Fahim Dalvi}, \bibinfo{person}{Nadir Durrani}, \bibinfo{person}{Hassan Sajjad}, \bibinfo{person}{Yonatan Belinkov}, \bibinfo{person}{Anthony Bau}, {and} \bibinfo{person}{James Glass}.} \bibinfo{year}{2019}\natexlab{}.
\newblock \showarticletitle{What is one grain of sand in the desert? Analyzing individual neurons in deep NLP models}.
\newblock \bibinfo{journal}{\emph{Proceedings of the AAAI Conference on Artificial Intelligence}} \bibinfo{volume}{33}, \bibinfo{number}{01} (\bibinfo{year}{2019}), \bibinfo{pages}{6309--6317}.
\newblock


\bibitem[de~Sousa et~al\mbox{.}(2021)]%
        {sousa2021explainable}
\bibfield{author}{\bibinfo{person}{Palatnik~I de Sousa}, \bibinfo{person}{MMBR Vellasco}, {and} \bibinfo{person}{Costa~E da Silva}.} \bibinfo{year}{2021}\natexlab{}.
\newblock \showarticletitle{Explainable Artificial Intelligence for Bias Detection in COVID CT-Scan Classifiers}.
\newblock \bibinfo{journal}{\emph{Sensors}} \bibinfo{volume}{21}, \bibinfo{number}{16} (\bibinfo{year}{2021}), \bibinfo{pages}{5657}.
\newblock


\bibitem[Gal and Ghahramani(2016)]%
        {gal2016dropout}
\bibfield{author}{\bibinfo{person}{Yarin Gal} {and} \bibinfo{person}{Zoubin Ghahramani}.} \bibinfo{year}{2016}\natexlab{}.
\newblock \showarticletitle{Dropout as a Bayesian approximation: Representing model uncertainty in deep learning}. In \bibinfo{booktitle}{\emph{ICML}}.
\newblock


\bibitem[Gallegos et~al\mbox{.}(2023)]%
        {gallegos2023bias}
\bibfield{author}{\bibinfo{person}{Isabel~O. Gallegos}, \bibinfo{person}{Ryan~A. Rossi}, \bibinfo{person}{Joe Barrow}, \bibinfo{person}{Md~Mehrab Tanjim}, \bibinfo{person}{Sungchul Kim}, \bibinfo{person}{Franck Dernoncourt}, \bibinfo{person}{Tong Yu}, \bibinfo{person}{Ruiyi Zhang}, {and} \bibinfo{person}{Nesreen~K. Ahmed}.} \bibinfo{year}{2023}\natexlab{}.
\newblock \showarticletitle{Bias and Fairness in Large Language Models: A Survey}.
\newblock \bibinfo{journal}{\emph{arXiv preprint arXiv:2309.00770}} (\bibinfo{year}{2023}).
\newblock


\bibitem[Gawlikowski et~al\mbox{.}(2023)]%
        {gawlikowski2023survey}
\bibfield{author}{\bibinfo{person}{Jakob Gawlikowski}, \bibinfo{person}{Cedrique~Rovile Njieutcheu~Tassi}, \bibinfo{person}{Mohsin Ali}, \bibinfo{person}{Jongseok Lee}, \bibinfo{person}{Matthias Humt}, \bibinfo{person}{Jianxing Feng}, \bibinfo{person}{Anna Kruspe}, \bibinfo{person}{Rudolph Triebel}, \bibinfo{person}{Peter Jung}, \bibinfo{person}{Ribana Roscher}, {et~al\mbox{.}}} \bibinfo{year}{2023}\natexlab{}.
\newblock \showarticletitle{A Survey of Uncertainty in Deep Neural Networks}.
\newblock \bibinfo{journal}{\emph{Artificial Intelligence Review}} (\bibinfo{year}{2023}).
\newblock


\bibitem[Guo and Caliskan(2021)]%
        {guo2021detecting}
\bibfield{author}{\bibinfo{person}{Wei Guo} {and} \bibinfo{person}{Aylin Caliskan}.} \bibinfo{year}{2021}\natexlab{}.
\newblock \showarticletitle{Detecting Emergent Intersectional Biases: Contextualized Word Embeddings Contain a Distribution of Human-like Biases}. In \bibinfo{booktitle}{\emph{Proceedings of the 2021 AAAI/ACM Conference on AI, Ethics, and Society}} (Virtual Event, USA) \emph{(\bibinfo{series}{AIES '21})}. \bibinfo{publisher}{Association for Computing Machinery}, \bibinfo{address}{New York, NY, USA}, \bibinfo{pages}{122–133}.
\newblock
\showISBNx{9781450384735}
\urldef\tempurl%
\url{https://doi.org/10.1145/3461702.3462536}
\showDOI{\tempurl}


\bibitem[Heuss et~al\mbox{.}(2023)]%
        {heuss2023predictive}
\bibfield{author}{\bibinfo{person}{Maria Heuss}, \bibinfo{person}{Daniel Cohen}, \bibinfo{person}{Masoud Mansoury}, \bibinfo{person}{Maarten de Rijke}, {and} \bibinfo{person}{Carsten Eickhoff}.} \bibinfo{year}{2023}\natexlab{}.
\newblock \showarticletitle{Predictive Uncertainty-based Bias Mitigation in Ranking}. In \bibinfo{booktitle}{\emph{Proceedings of the 32nd ACM International Conference on Information and Knowledge Management (CIKM '23)}}.
\newblock


\bibitem[Hort et~al\mbox{.}(2023)]%
        {hort2023bias}
\bibfield{author}{\bibinfo{person}{Max Hort}, \bibinfo{person}{Zhenpeng Chen}, \bibinfo{person}{Jie~M. Zhang}, \bibinfo{person}{Mark Harman}, {and} \bibinfo{person}{Federica Sarro}.} \bibinfo{year}{2023}\natexlab{}.
\newblock \bibinfo{title}{Bias Mitigation for Machine Learning Classifiers: A Comprehensive Survey}.
\newblock
\newblock
\showeprint[arxiv]{2207.07068}~[cs.LG]


\bibitem[Jain and Wallace(2019)]%
        {jain2019attention}
\bibfield{author}{\bibinfo{person}{Sarthak Jain} {and} \bibinfo{person}{Byron~C. Wallace}.} \bibinfo{year}{2019}\natexlab{}.
\newblock \showarticletitle{Attention is not explanation}.
\newblock \bibinfo{journal}{\emph{arXiv preprint arXiv:1902.10186}} (\bibinfo{year}{2019}).
\newblock


\bibitem[Jin et~al\mbox{.}(2020)]%
        {jin2020bert}
\bibfield{author}{\bibinfo{person}{Di Jin}, \bibinfo{person}{Zhijing Jin}, \bibinfo{person}{Joey~Tianyi Zhou}, {and} \bibinfo{person}{Peter Szolovits}.} \bibinfo{year}{2020}\natexlab{}.
\newblock \showarticletitle{Is BERT really robust? natural language attack on text classification and entailment}. In \bibinfo{booktitle}{\emph{AAAI Conference on Artificial Intelligence (AAAI)}}.
\newblock


\bibitem[Kim et~al\mbox{.}(2018)]%
        {kim2018interpretability}
\bibfield{author}{\bibinfo{person}{Been Kim}, \bibinfo{person}{Martin Wattenberg}, \bibinfo{person}{Justin Gilmer}, \bibinfo{person}{Carrie Cai}, \bibinfo{person}{James Wexler}, \bibinfo{person}{Fernanda Viegas}, {and} \bibinfo{person}{Rory Sayres}.} \bibinfo{year}{2018}\natexlab{}.
\newblock \showarticletitle{Interpretability beyond feature attribution: Quantitative testing with concept activation vectors (TCAV)}. In \bibinfo{booktitle}{\emph{Proceedings of the 35th International Conference on Machine Learning}}. PMLR, \bibinfo{pages}{2668--2677}.
\newblock


\bibitem[Kim et~al\mbox{.}(2020)]%
        {kim2020learning}
\bibfield{author}{\bibinfo{person}{Ildoo Kim}, \bibinfo{person}{Younghoon Kim}, {and} \bibinfo{person}{Sungwoong Kim}.} \bibinfo{year}{2020}\natexlab{}.
\newblock \showarticletitle{Learning loss for test-time augmentation}. In \bibinfo{booktitle}{\emph{Proceedings of the 34th International Conference on Neural Information Processing Systems}} (Vancouver, BC, Canada) \emph{(\bibinfo{series}{NIPS'20})}. \bibinfo{publisher}{Curran Associates Inc.}, \bibinfo{address}{Red Hook, NY, USA}, Article \bibinfo{articleno}{350}, \bibinfo{numpages}{12}~pages.
\newblock
\showISBNx{9781713829546}


\bibitem[Kuleshov et~al\mbox{.}(2018)]%
        {kuleshov2018accurate}
\bibfield{author}{\bibinfo{person}{Volodymyr Kuleshov}, \bibinfo{person}{Nathan Fenner}, {and} \bibinfo{person}{Stefano Ermon}.} \bibinfo{year}{2018}\natexlab{}.
\newblock \showarticletitle{Accurate Uncertainties for Deep Learning Using Calibrated Regression}. In \bibinfo{booktitle}{\emph{Proceedings of the 35th International Conference on Machine Learning}} \emph{(\bibinfo{series}{Proceedings of Machine Learning Research}, Vol.~\bibinfo{volume}{80})}, \bibfield{editor}{\bibinfo{person}{Jennifer Dy} {and} \bibinfo{person}{Andreas Krause}} (Eds.). \bibinfo{publisher}{PMLR}, \bibinfo{pages}{2796--2804}.
\newblock
\urldef\tempurl%
\url{https://proceedings.mlr.press/v80/kuleshov18a.html}
\showURL{%
\tempurl}


\bibitem[Kunz and Kuhlmann(2020)]%
        {kunz2020classifier}
\bibfield{author}{\bibinfo{person}{Jenny Kunz} {and} \bibinfo{person}{Marco Kuhlmann}.} \bibinfo{year}{2020}\natexlab{}.
\newblock \showarticletitle{Classifier probes may just learn from linear context features}. In \bibinfo{booktitle}{\emph{Proceedings of the 28th International Conference on Computational Linguistics}}. International Committee on Computational Linguistics, \bibinfo{pages}{5136--5146}.
\newblock


\bibitem[Lakshminarayanan et~al\mbox{.}(2017)]%
        {lakshminarayanan2017simple}
\bibfield{author}{\bibinfo{person}{Balaji Lakshminarayanan}, \bibinfo{person}{Alexander Pritzel}, {and} \bibinfo{person}{Charles Blundell}.} \bibinfo{year}{2017}\natexlab{}.
\newblock \showarticletitle{Simple and scalable predictive uncertainty estimation using deep ensembles}. In \bibinfo{booktitle}{\emph{Advances in Neural Information Processing Systems}}, Vol.~\bibinfo{volume}{30}.
\newblock


\bibitem[Li et~al\mbox{.}(2016)]%
        {li2016visualizing}
\bibfield{author}{\bibinfo{person}{Jiwei Li}, \bibinfo{person}{Xinlei Chen}, \bibinfo{person}{Eduard Hovy}, {and} \bibinfo{person}{Dan Jurafsky}.} \bibinfo{year}{2016}\natexlab{}.
\newblock \showarticletitle{Visualizing and Understanding Neural Models in NLP}. In \bibinfo{booktitle}{\emph{Proceedings of the 2016 Conference of the North American Chapter of the Association for Computational Linguistics: Human Language Technologies}}. \bibinfo{address}{San Diego, California}, \bibinfo{pages}{681--691}.
\newblock


\bibitem[Li et~al\mbox{.}(2023)]%
        {li2023survey}
\bibfield{author}{\bibinfo{person}{Yingji Li}, \bibinfo{person}{Mengnan Du}, \bibinfo{person}{Rui Song}, \bibinfo{person}{Xin Wang}, {and} \bibinfo{person}{Ying Wang}.} \bibinfo{year}{2023}\natexlab{}.
\newblock \bibinfo{title}{A Survey on Fairness in Large Language Models}.
\newblock
\newblock
\showeprint[arxiv]{2308.10149}~[cs.CL]


\bibitem[Liu et~al\mbox{.}(2023)]%
        {li2023towards}
\bibfield{author}{\bibinfo{person}{Yibing Liu}, \bibinfo{person}{Haoliang Li}, \bibinfo{person}{Yangyang Guo}, \bibinfo{person}{Chenqi Kong}, \bibinfo{person}{Jing Li}, {and} \bibinfo{person}{Shiqi Wang}.} \bibinfo{year}{2023}\natexlab{}.
\newblock \showarticletitle{Towards understanding in-context learning with contrastive demonstrations and saliency maps}.
\newblock \bibinfo{journal}{\emph{arXiv preprint arXiv:2308.10149}} (\bibinfo{year}{2023}).
\newblock


\bibitem[Lundberg and Lee(2017)]%
        {lundberg2017unified}
\bibfield{author}{\bibinfo{person}{Scott~M. Lundberg} {and} \bibinfo{person}{Su-In Lee}.} \bibinfo{year}{2017}\natexlab{}.
\newblock \showarticletitle{A unified approach to interpreting model predictions}. In \bibinfo{booktitle}{\emph{Advances in Neural Information Processing Systems}}, Vol.~\bibinfo{volume}{30}. Curran Associates, Inc.
\newblock


\bibitem[Madaan and Yazdanbakhsh(2022)]%
        {madaan2022text}
\bibfield{author}{\bibinfo{person}{Aman Madaan} {and} \bibinfo{person}{Amir Yazdanbakhsh}.} \bibinfo{year}{2022}\natexlab{}.
\newblock \showarticletitle{Text and patterns: For effective chain of thought, it takes two to tango}.
\newblock \bibinfo{journal}{\emph{arXiv preprint arXiv:2209.07686}} (\bibinfo{year}{2022}).
\newblock


\bibitem[Maudslay and Cotterell(2021)]%
        {maudslay2021do}
\bibfield{author}{\bibinfo{person}{Rowan~Hall Maudslay} {and} \bibinfo{person}{Ryan Cotterell}.} \bibinfo{year}{2021}\natexlab{}.
\newblock \showarticletitle{Do syntactic probes probe syntax? experiments with jabberwocky probing}. In \bibinfo{booktitle}{\emph{Proceedings of the 2021 Conference of the North American Chapter of the Association for Computational Linguistics: Human Language Technologies}}. Association for Computational Linguistics, \bibinfo{pages}{144--150}.
\newblock


\bibitem[May et~al\mbox{.}(2019)]%
        {may2019measuring}
\bibfield{author}{\bibinfo{person}{Chandler May}, \bibinfo{person}{Alex Wang}, \bibinfo{person}{Shikha Bordia}, \bibinfo{person}{Samuel~R Bowman}, {and} \bibinfo{person}{Rachel Rudinger}.} \bibinfo{year}{2019}\natexlab{}.
\newblock \showarticletitle{On measuring social biases in sentence encoders}. In \bibinfo{booktitle}{\emph{Proceedings of the 2019 Conference of the North American Chapter of the Association for Computational Linguistics: Human Language Technologies, Volume 1 (Long and Short Papers)}}. Association for Computational Linguistics, \bibinfo{pages}{622--628}.
\newblock
\urldef\tempurl%
\url{https://doi.org/10.18653/v1/N19-1063}
\showDOI{\tempurl}


\bibitem[Mehrabi et~al\mbox{.}(2022)]%
        {mehrabi2022survey}
\bibfield{author}{\bibinfo{person}{Ninareh Mehrabi}, \bibinfo{person}{Fred Morstatter}, \bibinfo{person}{Nripsuta Saxena}, \bibinfo{person}{Kristina Lerman}, {and} \bibinfo{person}{Aram Galstyan}.} \bibinfo{year}{2022}\natexlab{}.
\newblock \showarticletitle{A Survey on Bias and Fairness in Machine Learning}.
\newblock \bibinfo{journal}{\emph{Comput. Surveys}} \bibinfo{volume}{54}, \bibinfo{number}{6} (\bibinfo{year}{2022}), \bibinfo{pages}{1--35}.
\newblock


\bibitem[Molchanov et~al\mbox{.}(2020)]%
        {molchanov2020greedy}
\bibfield{author}{\bibinfo{person}{Dmitry Molchanov}, \bibinfo{person}{Alexander Lyzhov}, \bibinfo{person}{Yuliya Molchanova}, \bibinfo{person}{Arsenii Ashukha}, {and} \bibinfo{person}{Dmitry Vetrov}.} \bibinfo{year}{2020}\natexlab{}.
\newblock \bibinfo{title}{Greedy Policy Search: A Simple Baseline for Learnable Test-Time Augmentation}.
\newblock
\newblock
\showeprint[arxiv]{2002.09103}~[stat.ML]


\bibitem[Montavon et~al\mbox{.}(2015)]%
        {montavon2015explaining}
\bibfield{author}{\bibinfo{person}{Gr{\'e}goire Montavon}, \bibinfo{person}{Sebastian Bach}, \bibinfo{person}{Alexander Binder}, \bibinfo{person}{Wojciech Samek}, {and} \bibinfo{person}{Klaus-Robert M{\"u}ller}.} \bibinfo{year}{2015}\natexlab{}.
\newblock \showarticletitle{Explaining NonLinear Classification Decisions with Deep Taylor Decomposition}.
\newblock \bibinfo{journal}{\emph{arXiv preprint arXiv:1512.02479}} (\bibinfo{year}{2015}).
\newblock


\bibitem[Montavon et~al\mbox{.}(2019)]%
        {montavon2019layer}
\bibfield{author}{\bibinfo{person}{Gr{\'e}goire Montavon}, \bibinfo{person}{Alexander Binder}, \bibinfo{person}{Sebastian Lapuschkin}, \bibinfo{person}{Wojciech Samek}, {and} \bibinfo{person}{Klaus-Robert M{\"u}ller}.} \bibinfo{year}{2019}\natexlab{}.
\newblock \showarticletitle{Layer-wise relevance propagation: an overview}.
\newblock \bibinfo{journal}{\emph{Explainable AI: interpreting, explaining and visualizing deep learning}} (\bibinfo{year}{2019}), \bibinfo{pages}{193--209}.
\newblock


\bibitem[Mouli et~al\mbox{.}(2022)]%
        {mouli2022bias}
\bibfield{author}{\bibinfo{person}{S~Chandra Mouli}, \bibinfo{person}{Yangze Zhou}, {and} \bibinfo{person}{Bruno Ribeiro}.} \bibinfo{year}{2022}\natexlab{}.
\newblock \bibinfo{title}{Bias Challenges in Counterfactual Data Augmentation}.
\newblock
\newblock
\showeprint[arxiv]{2209.05104}~[cs.LG]


\bibitem[Nadeem et~al\mbox{.}(2021)]%
        {nadeem2021stereoset}
\bibfield{author}{\bibinfo{person}{Moin Nadeem}, \bibinfo{person}{Anna Bethke}, {and} \bibinfo{person}{Siva Reddy}.} \bibinfo{year}{2021}\natexlab{}.
\newblock \showarticletitle{StereoSet: Measuring stereotypical bias in pretrained language models}. In \bibinfo{booktitle}{\emph{ACL}}.
\newblock


\bibitem[N{\'a}poles and Koutsoviti~Koumeri(2022)]%
        {napoles2022fuzzyrough}
\bibfield{author}{\bibinfo{person}{Gonzalo N{\'a}poles} {and} \bibinfo{person}{Lisa Koutsoviti~Koumeri}.} \bibinfo{year}{2022}\natexlab{}.
\newblock \showarticletitle{A fuzzy-rough uncertainty measure to discover bias encoded explicitly or implicitly in features of structured pattern classification datasets}.
\newblock \bibinfo{journal}{\emph{Pattern Recognition Letters}}  \bibinfo{volume}{154} (\bibinfo{year}{2022}).
\newblock


\bibitem[Nguyen et~al\mbox{.}(2023)]%
        {nguyen}
\bibfield{author}{\bibinfo{person}{Thanh-Tung Nguyen}, \bibinfo{person}{Viktor Schlegel}, \bibinfo{person}{Abhinav Kashyap}, \bibinfo{person}{Stefan Winkler}, \bibinfo{person}{Shao-Syuan Huang}, \bibinfo{person}{Jie-Jyun Liu}, {and} \bibinfo{person}{Chih-Jen Lin}.} \bibinfo{year}{2023}\natexlab{}.
\newblock \bibinfo{title}{Mimic-IV-ICD: A new benchmark for eXtreme MultiLabel Classification}.
\newblock
\newblock
\showeprint[arxiv]{2304.13998}~[cs.AI]


\bibitem[Nozza et~al\mbox{.}(2021)]%
        {nozza2021honest}
\bibfield{author}{\bibinfo{person}{Debora Nozza}, \bibinfo{person}{Federico Bianchi}, {and} \bibinfo{person}{Dirk Hovy}.} \bibinfo{year}{2021}\natexlab{}.
\newblock \showarticletitle{HONEST: Measuring hurtful sentence completion in language models}. In \bibinfo{booktitle}{\emph{Proceedings of the 2021 Conference of the North American Chapter of the Association for Computational Linguistics: Human Language Technologies}}. Association for Computational Linguistics, \bibinfo{pages}{2398--2406}.
\newblock
\urldef\tempurl%
\url{https://doi.org/10.18653/v1/2021.naacl-main.191}
\showDOI{\tempurl}


\bibitem[Omiye et~al\mbox{.}(2023)]%
        {omiye2023large}
\bibfield{author}{\bibinfo{person}{Jesutofunmi~A. Omiye}, \bibinfo{person}{Jenna~C. Lester}, \bibinfo{person}{Simon Spichak}, \bibinfo{person}{Veronica Rotemberg}, {and} \bibinfo{person}{Roxana Daneshjou}.} \bibinfo{year}{2023}\natexlab{}.
\newblock \showarticletitle{Large language models propagate race-based medicine}.
\newblock \bibinfo{journal}{\emph{npj Digital Medicine}} \bibinfo{volume}{6}, \bibinfo{number}{195} (\bibinfo{year}{2023}).
\newblock
\urldef\tempurl%
\url{https://doi.org/10.1038/s41746-023-00939-z}
\showDOI{\tempurl}


\bibitem[OpenAI(2023a)]%
        {openai2023gpt4}
\bibfield{author}{\bibinfo{person}{OpenAI}.} \bibinfo{year}{2023}\natexlab{a}.
\newblock \bibinfo{title}{GPT-4 Technical Report}.
\newblock
\newblock
\showeprint[arxiv]{2303.08774}~[cs.CL]


\bibitem[OpenAI(2023b)]%
        {openai2023language}
\bibfield{author}{\bibinfo{person}{OpenAI}.} \bibinfo{year}{2023}\natexlab{b}.
\newblock \bibinfo{title}{Language models can explain neurons in language models}.
\newblock
\newblock
\urldef\tempurl%
\url{https://openai.com/research/language-models-can-explain-neurons-in-language-models}
\showURL{%
\tempurl}


\bibitem[Raffel et~al\mbox{.}(2020)]%
        {raffel2020exploring}
\bibfield{author}{\bibinfo{person}{Colin Raffel}, \bibinfo{person}{Noam Shazeer}, \bibinfo{person}{Adam Roberts}, \bibinfo{person}{Katherine Lee}, \bibinfo{person}{Sharan Narang}, \bibinfo{person}{Michael Matena}, \bibinfo{person}{Yanqi Zhou}, \bibinfo{person}{Wei Li}, {and} \bibinfo{person}{Peter~J. Liu}.} \bibinfo{year}{2020}\natexlab{}.
\newblock \showarticletitle{Exploring the limits of transfer learning with a unified text-to-text transformer}.
\newblock \bibinfo{journal}{\emph{Journal of Machine Learning Research}} \bibinfo{volume}{21}, \bibinfo{number}{1} (\bibinfo{year}{2020}), \bibinfo{pages}{5485--5551}.
\newblock


\bibitem[Ribeiro et~al\mbox{.}(2016a)]%
        {ribeiro2016should}
\bibfield{author}{\bibinfo{person}{Marco~Tulio Ribeiro}, \bibinfo{person}{Sameer Singh}, {and} \bibinfo{person}{Carlos Guestrin}.} \bibinfo{year}{2016}\natexlab{a}.
\newblock \showarticletitle{"Why should I trust you?": Explaining the predictions of any classifier}. In \bibinfo{booktitle}{\emph{Proceedings of the 22nd ACM SIGKDD International Conference on Knowledge Discovery and Data Mining}}. ACM, \bibinfo{pages}{1135--1144}.
\newblock


\bibitem[Ribeiro et~al\mbox{.}(2016b)]%
        {ribeiro2016why}
\bibfield{author}{\bibinfo{person}{Marco~Tulio Ribeiro}, \bibinfo{person}{Sameer Singh}, {and} \bibinfo{person}{Carlos Guestrin}.} \bibinfo{year}{2016}\natexlab{b}.
\newblock \showarticletitle{"Why Should I Trust You?": Explaining the Predictions of Any Classifier}. In \bibinfo{booktitle}{\emph{Proceedings of the 22nd ACM SIGKDD International Conference on Knowledge Discovery and Data Mining (KDD '16)}}.
\newblock


\bibitem[Sensoy et~al\mbox{.}(2018)]%
        {sensoy2018evidential}
\bibfield{author}{\bibinfo{person}{Murat Sensoy}, \bibinfo{person}{Lance Kaplan}, {and} \bibinfo{person}{Melih Kandemir}.} \bibinfo{year}{2018}\natexlab{}.
\newblock \showarticletitle{Evidential Deep Learning to Quantify Classification Uncertainty}. In \bibinfo{booktitle}{\emph{Advances in Neural Information Processing Systems}}, \bibfield{editor}{\bibinfo{person}{S.~Bengio}, \bibinfo{person}{H.~Wallach}, \bibinfo{person}{H.~Larochelle}, \bibinfo{person}{K.~Grauman}, \bibinfo{person}{N.~Cesa-Bianchi}, {and} \bibinfo{person}{R.~Garnett}} (Eds.), Vol.~\bibinfo{volume}{31}. \bibinfo{publisher}{Curran Associates, Inc.}
\newblock
\urldef\tempurl%
\url{https://proceedings.neurips.cc/paper_files/paper/2018/file/a981f2b708044d6fb4a71a1463242520-Paper.pdf}
\showURL{%
\tempurl}


\bibitem[Shanmugam et~al\mbox{.}(2020)]%
        {shanmugam2020testtime}
\bibfield{author}{\bibinfo{person}{D. Shanmugam}, \bibinfo{person}{D. Blalock}, \bibinfo{person}{G. Balakrishnan}, {and} \bibinfo{person}{J. Guttag}.} \bibinfo{year}{2020}\natexlab{}.
\newblock \showarticletitle{When and why test-time augmentation works}.
\newblock \bibinfo{journal}{\emph{arXiv preprint arXiv:2011.11156}} (\bibinfo{year}{2020}).
\newblock


\bibitem[Shorten et~al\mbox{.}(2021)]%
        {Shorten2021TextDataAugmentation}
\bibfield{author}{\bibinfo{person}{Conner Shorten}, \bibinfo{person}{Taghi~M. Khoshgoftaar}, {and} \bibinfo{person}{Borko Furht}.} \bibinfo{year}{2021}\natexlab{}.
\newblock \showarticletitle{Text Data Augmentation for Deep Learning}.
\newblock \bibinfo{journal}{\emph{Journal of Big Data}}  \bibinfo{volume}{8} (\bibinfo{year}{2021}), \bibinfo{pages}{101}.
\newblock
\urldef\tempurl%
\url{https://doi.org/10.1186/s40537-021-00492-0}
\showDOI{\tempurl}


\bibitem[Singh and Ramakrishnan(2023)]%
        {singh2023chatgptbiased}
\bibfield{author}{\bibinfo{person}{Sahib Singh} {and} \bibinfo{person}{Narayanan Ramakrishnan}.} \bibinfo{year}{2023}\natexlab{}.
\newblock \bibinfo{booktitle}{\emph{{Is ChatGPT Biased? A Review}}}.
\newblock
\urldef\tempurl%
\url{https://osf.io/preprints/your-preprint-id}
\showURL{%
\tempurl}
\newblock
\shownote{Web}.


\bibitem[Stone et~al\mbox{.}(2023)]%
        {stone2023implicit}
\bibfield{author}{\bibinfo{person}{Rebecca~S Stone}, \bibinfo{person}{Nishant Ravikumar}, \bibinfo{person}{Andrew~J Bulpitt}, {and} \bibinfo{person}{David~C Hogg}.} \bibinfo{year}{2023}\natexlab{}.
\newblock \showarticletitle{Implicit Visual Bias Mitigation by Posterior Estimate Sharpening of a Bayesian Neural Network}.
\newblock \bibinfo{journal}{\emph{arXiv preprint arXiv:2303.16564}} (\bibinfo{year}{2023}).
\newblock


\bibitem[Suresh and Guttag(2021)]%
        {suresh2021framework}
\bibfield{author}{\bibinfo{person}{Harini Suresh} {and} \bibinfo{person}{John Guttag}.} \bibinfo{year}{2021}\natexlab{}.
\newblock \showarticletitle{A framework for understanding sources of harm throughout the machine learning life cycle}.
\newblock \bibinfo{journal}{\emph{Equity and Access in Algorithms, Mechanisms, and Optimization}} (\bibinfo{year}{2021}), \bibinfo{pages}{1--9}.
\newblock


\bibitem[Urchs et~al\mbox{.}(2023)]%
        {urchs2023prevalent}
\bibfield{author}{\bibinfo{person}{Stefanie Urchs}, \bibinfo{person}{Veronika Thurner}, \bibinfo{person}{Matthias A{\ss}enmacher}, \bibinfo{person}{Christian Heumann}, {and} \bibinfo{person}{Stephanie Thiemichen}.} \bibinfo{year}{2023}\natexlab{}.
\newblock \showarticletitle{How Prevalent is Gender Bias in ChatGPT? - Exploring German and English ChatGPT Responses}. In \bibinfo{booktitle}{\emph{1st Workshop onBiased Data in Conversational Agents - colocated with ECML-PKDD '23}}.
\newblock


\bibitem[Wachter et~al\mbox{.}(2017)]%
        {wachter2017counterfactual}
\bibfield{author}{\bibinfo{person}{Sandra Wachter}, \bibinfo{person}{Brent Mittelstadt}, {and} \bibinfo{person}{Chris Russell}.} \bibinfo{year}{2017}\natexlab{}.
\newblock \showarticletitle{Counterfactual explanations without opening the black box: Automated decisions and the GDPR}.
\newblock \bibinfo{journal}{\emph{Harv. JL \& Tech.}}  \bibinfo{volume}{31} (\bibinfo{year}{2017}), \bibinfo{pages}{841}.
\newblock


\bibitem[Wan et~al\mbox{.}(2023)]%
        {wan2023kelly}
\bibfield{author}{\bibinfo{person}{Yixin Wan}, \bibinfo{person}{George Pu}, \bibinfo{person}{Jiao Sun}, \bibinfo{person}{Aparna Garimella}, \bibinfo{person}{Kai-Wei Chang}, {and} \bibinfo{person}{Nanyun Peng}.} \bibinfo{year}{2023}\natexlab{}.
\newblock \bibinfo{title}{"Kelly is a Warm Person, Joseph is a Role Model": Gender Biases in LLM-Generated Reference Letters}.
\newblock
\newblock
\showeprint[arxiv]{2310.09219}~[cs.CL]


\bibitem[Wang et~al\mbox{.}(2022a)]%
        {wang2022towards}
\bibfield{author}{\bibinfo{person}{Boshi Wang}, \bibinfo{person}{Sewon Min}, \bibinfo{person}{Xiang Deng}, \bibinfo{person}{Jiaming Shen}, \bibinfo{person}{You Wu}, \bibinfo{person}{Luke Zettlemoyer}, {and} \bibinfo{person}{Huan Sun}.} \bibinfo{year}{2022}\natexlab{a}.
\newblock \showarticletitle{Towards understanding chain-of-thought prompting: An empirical study of what matters}.
\newblock \bibinfo{journal}{\emph{arXiv preprint arXiv:2212.10001}} (\bibinfo{year}{2022}).
\newblock


\bibitem[Wang et~al\mbox{.}(2022b)]%
        {wang2022selfconsistency}
\bibfield{author}{\bibinfo{person}{Xuezhi Wang}, \bibinfo{person}{Jason Wei}, \bibinfo{person}{Dale Schuurmans}, \bibinfo{person}{Quoc Le}, \bibinfo{person}{Ed Chi}, {and} \bibinfo{person}{Denny Zhou}.} \bibinfo{year}{2022}\natexlab{b}.
\newblock \showarticletitle{Self-consistency improves chain of thought reasoning in language models}.
\newblock \bibinfo{journal}{\emph{arXiv preprint arXiv:2203.11171}} (\bibinfo{year}{2022}).
\newblock


\bibitem[Webster et~al\mbox{.}(2020)]%
        {webster2020measuring}
\bibfield{author}{\bibinfo{person}{Kellie Webster}, \bibinfo{person}{Xuezhi Wang}, \bibinfo{person}{Ian Tenney}, \bibinfo{person}{Alex Beutel}, \bibinfo{person}{Emily Pitler}, \bibinfo{person}{Ellie Pavlick}, \bibinfo{person}{Jilin Chen}, \bibinfo{person}{Ed Chi}, {and} \bibinfo{person}{Slav Petrov}.} \bibinfo{year}{2020}\natexlab{}.
\newblock \showarticletitle{Measuring and reducing gendered correlations in pre-trained models}.
\newblock \bibinfo{journal}{\emph{arXiv preprint arXiv:2010.06032}} (\bibinfo{year}{2020}).
\newblock
\urldef\tempurl%
\url{https://arxiv.org/abs/2010.06032}
\showURL{%
\tempurl}


\bibitem[Wei et~al\mbox{.}(2022)]%
        {wei2022chain}
\bibfield{author}{\bibinfo{person}{Jerry Wei}, \bibinfo{person}{Xuezhi Wang}, \bibinfo{person}{Dale Schuurmans}, \bibinfo{person}{Maarten Bosma}, \bibinfo{person}{Fei Xia}, \bibinfo{person}{Ed Chi}, \bibinfo{person}{Quoc~V Le}, {and} \bibinfo{person}{Denny Zhou}.} \bibinfo{year}{2022}\natexlab{}.
\newblock \showarticletitle{Chain-of-thought prompting elicits reasoning in large language models}.
\newblock \bibinfo{journal}{\emph{Advances in Neural Information Processing Systems}}  \bibinfo{volume}{35} (\bibinfo{year}{2022}), \bibinfo{pages}{24824--24837}.
\newblock


\bibitem[Wei et~al\mbox{.}(2023)]%
        {wei2023larger}
\bibfield{author}{\bibinfo{person}{Jerry Wei}, \bibinfo{person}{Jason Wei}, \bibinfo{person}{Yi Tay}, \bibinfo{person}{Dustin Tran}, \bibinfo{person}{Albert Webson}, \bibinfo{person}{Yifeng Lu}, \bibinfo{person}{Xinyun Chen}, \bibinfo{person}{Hanxiao Liu}, \bibinfo{person}{Da Huang}, {and} \bibinfo{person}{Denny Zhou}.} \bibinfo{year}{2023}\natexlab{}.
\newblock \showarticletitle{Larger language models do in-context learning differently}.
\newblock \bibinfo{journal}{\emph{arXiv preprint arXiv:2303.03846}} (\bibinfo{year}{2023}).
\newblock


\bibitem[Wu et~al\mbox{.}(2023)]%
        {wu2023analyzing}
\bibfield{author}{\bibinfo{person}{Skyler Wu}, \bibinfo{person}{Eric~Meng Shen}, \bibinfo{person}{Charumathi Badrinath}, \bibinfo{person}{Jiaqi Ma}, {and} \bibinfo{person}{Himabindu Lakkaraju}.} \bibinfo{year}{2023}\natexlab{}.
\newblock \showarticletitle{Analyzing chain-of-thought prompting in large language models via gradient-based feature attributions}.
\newblock \bibinfo{journal}{\emph{arXiv preprint arXiv:2306.01150}} (\bibinfo{year}{2023}).
\newblock


\bibitem[Wu et~al\mbox{.}(2021)]%
        {wu2021polyjuice}
\bibfield{author}{\bibinfo{person}{Tongshuang Wu}, \bibinfo{person}{Marco~Tulio Ribeiro}, \bibinfo{person}{Jeffrey Heer}, {and} \bibinfo{person}{Daniel Weld}.} \bibinfo{year}{2021}\natexlab{}.
\newblock \showarticletitle{Polyjuice: Generating Counterfactuals for Explaining, Evaluating, and Improving Models}. In \bibinfo{booktitle}{\emph{Proceedings of the 59th Annual Meeting of the Association for Computational Linguistics and the 11th International Joint Conference on Natural Language Processing (Volume 1: Long Papers)}}. Online, \bibinfo{pages}{6707--6723}.
\newblock


\bibitem[Wu et~al\mbox{.}(2020)]%
        {wu2020perturbed}
\bibfield{author}{\bibinfo{person}{Zhiyong Wu}, \bibinfo{person}{Yun Chen}, \bibinfo{person}{Ben Kao}, {and} \bibinfo{person}{Qun Liu}.} \bibinfo{year}{2020}\natexlab{}.
\newblock \showarticletitle{Perturbed Masking: Parameter-free Probing for Analyzing and Interpreting BERT}. In \bibinfo{booktitle}{\emph{Proceedings of the 58th Annual Meeting of the Association for Computational Linguistics}}. Online, \bibinfo{pages}{4166--4176}.
\newblock


\bibitem[Xiong et~al\mbox{.}(2023)]%
        {xiong2023can}
\bibfield{author}{\bibinfo{person}{Miao Xiong}, \bibinfo{person}{Zhiyuan Hu}, \bibinfo{person}{Xinyang Lu}, \bibinfo{person}{Yifei Li}, \bibinfo{person}{Jie Fu}, \bibinfo{person}{Junxian He}, {and} \bibinfo{person}{Bryan Hooi}.} \bibinfo{year}{2023}\natexlab{}.
\newblock \showarticletitle{Can LLMs Express Their Uncertainty? An Empirical Evaluation of Confidence Elicitation in LLMs}.
\newblock \bibinfo{journal}{\emph{arXiv preprint arXiv:2306.13063}} (\bibinfo{year}{2023}).
\newblock


\bibitem[Zhao et~al\mbox{.}(2023)]%
        {zhao2023explainability}
\bibfield{author}{\bibinfo{person}{Haiyan Zhao}, \bibinfo{person}{Hanjie Chen}, \bibinfo{person}{Fan Yang}, \bibinfo{person}{Ninghao Liu}, \bibinfo{person}{Huiqi Deng}, \bibinfo{person}{Hengyi Cai}, \bibinfo{person}{Shuaiqiang Wang}, \bibinfo{person}{Dawei Yin}, {and} \bibinfo{person}{Mengnan Du}.} \bibinfo{year}{2023}\natexlab{}.
\newblock \showarticletitle{Explainability for Large Language Models: A Survey}.
\newblock \bibinfo{journal}{\emph{arXiv preprint arXiv:2309.01029}} (\bibinfo{year}{2023}).
\newblock


\bibitem[Zhou et~al\mbox{.}(2023)]%
        {zhou2023lima}
\bibfield{author}{\bibinfo{person}{Chunting Zhou}, \bibinfo{person}{Pengfei Liu}, \bibinfo{person}{Puxin Xu}, \bibinfo{person}{Srini Iyer}, \bibinfo{person}{Jiao Sun}, \bibinfo{person}{Yuning Mao}, \bibinfo{person}{Xuezhe Ma}, \bibinfo{person}{Avia Efrat}, \bibinfo{person}{Ping Yu}, {and} \bibinfo{person}{Lili Yu}.} \bibinfo{year}{2023}\natexlab{}.
\newblock \showarticletitle{Lima: Less is more for alignment}.
\newblock \bibinfo{journal}{\emph{arXiv preprint arXiv:2305.11206}} (\bibinfo{year}{2023}).
\newblock


\bibitem[Zmigrod et~al\mbox{.}(2019)]%
        {Zmigrod2019CounterfactualDA}
\bibfield{author}{\bibinfo{person}{Ran Zmigrod}, \bibinfo{person}{Sabrina~J. Mielke}, \bibinfo{person}{Hanna~M. Wallach}, {and} \bibinfo{person}{Ryan Cotterell}.} \bibinfo{year}{2019}\natexlab{}.
\newblock \showarticletitle{Counterfactual Data Augmentation for Mitigating Gender Stereotypes in Languages with Rich Morphology}.
\newblock \bibinfo{journal}{\emph{ArXiv}}  \bibinfo{volume}{abs/1906.04571} (\bibinfo{year}{2019}).
\newblock
\urldef\tempurl%
\url{https://api.semanticscholar.org/CorpusID:184486914}
\showURL{%
\tempurl}


\end{thebibliography}

\end{document}